\documentclass[1p]{elsarticle}

\usepackage{hyperref}

\journal{ }

\usepackage[utf8]{inputenc} 
\usepackage[T1]{fontenc}    
\usepackage{hyperref}       
\usepackage{url}            
\usepackage{booktabs}       
\usepackage{amsfonts}       
\usepackage{nicefrac}       
\usepackage{microtype}      

\usepackage{times}
\usepackage{epsfig}
\usepackage{graphicx}
\usepackage{amsmath}
\usepackage{amssymb}
\usepackage{epstopdf} 
\usepackage{marvosym}

\usepackage{subfigure}
\usepackage{overpic}
\usepackage{multirow}

\begin{document}

\begin{frontmatter}

\title{Re-ranking Object Proposals for Object Detection in Automatic Driving}


\author{Zhun Zhong\fnref{myfootnote}}
\fntext[myfootnote]{Cognitive Science Department, Xiamen University,  Xiamen,  Fujian, 361005, China}

\author{Mingyi Lei\fnref{myfootnote}}

\author[mysecondaryaddress]{Shaozi Li}

\author{Jianping Fan \fnref{myfootnote1}}
\fntext[myfootnote1]{Department of Computer Science, UNC-Charlotte, Charlotte, NC 28223, USA}

\begin{abstract}
Object detection often suffers from a plenty of bootless proposals, 
	selecting high quality proposals remains a great challenge.
In this paper, we propose a semantic, class-specific approach to re-rank object proposals, 
	which can consistently improve the recall performance even with less proposals.
We first extract features for each proposal including semantic segmentation, stereo information, contextual information, CNN-based objectness and low-level cue,
	and then score them using class-specific weights learnt by Structured SVM.
The advantages of the proposed model are two-fold: 
 	1) it can be easily merged to existing generators with few computational costs,
	and 2) it can achieve high recall rate uner strict critical even using less proposals.
Experimental evaluation on the KITTI benchmark demonstrates that our approach significantly improves existing popular generators on recall performance.
Moreover, in the experiment conducted for object detection, 
	even with 1,500 proposals, our approach can still have higher average precision (AP) than baselines with 5,000 proposals.
\end{abstract}

\begin{keyword}
Re-ranking\sep Object proposal\sep Object detection \sep CNN
\end{keyword}

\end{frontmatter}


\section{Introduction}
In the last few years, object proposal methods have been  successfully applied to a number of computer vision tasks, 
	such as object detection \cite{girshick2014rich-RCNN, girshick2015fastrcnn}, object segmentation \cite{dai2015convolutional}, and object discovery \cite{cho2015unsupervised}. 
Especially in object detection, object proposal methods have achieved great success.
The goal of object proposal methods is to generate a set of candidate regions in an image that are likely to contain objects. 
In contrast to sliding window paradigm \cite{papageorgiou2000trainable}, object proposal methods generate fewer candidate regions by reducing the search space, 
	which significantly reduces computation cost for subsequent detection process, 
	and enables the usage of more sophisticated classifier to obtain more accurate results. 
In addition, object proposal methods can make detection easier by removing false positives \cite{hosang2015makes}.

Most existing state-of-the-art object proposal methods mainly depend on bottom-up grouping and saliency cues to generate and rank proposals. 
They commonly aim to generate class-agnostic proposals in a reasonable time consumption. 
These object proposals methods have already been proven to achieve high recall performance and satisfactory detection accuracy in the popular ILSVRC \cite{russakovsky2015imagenet} and PASCAL VOC \cite{everingham2015pascal} Detection Challenge Benchmark, 
	which require loose criteria, i.e. a detection is regarded as correct if the intersection over union (IoU) overlap is more than 0.5.
However, these object proposal methods fail under strict criteria (e.g IoU > 0.7) even if  the state-of-the-art R-CNN \cite{girshick2014rich-RCNN} object detection approach is employed.
Especially in the considerably challenge KITTI \cite{geiger2012KITTI} benchmark, 
	their performance is barely satisfactory since only low-level cues are considered.

More recently, DeepBox \cite{kuo2015deepbox} proposes a CNN-based object proposal re-ranking method, 
	which exploits the high-level structures to compute the objectness of candidate proposals, and re-ranks proposals using the computed objectness.
Similarly, RPN \cite{ren2015fasterrcnn} proposes a method scores the proposals based on the objectness of CNN network. 
These methods have achieved high recall rate with loose criteria, 			however, strict criteria still bring a big challenge to them.

Nearly all of mentioned methods, low-level cues based or high-level cues based, 
	adopt class-agnostic scoring strategy and struggle to achieve high recall under strict criteria. 
This motivates us to improve the object proposals recall across various IoU thresholds (especially strict criteria).

\begin{figure}[t]
    \centering
    \begin{overpic}[width=\linewidth]{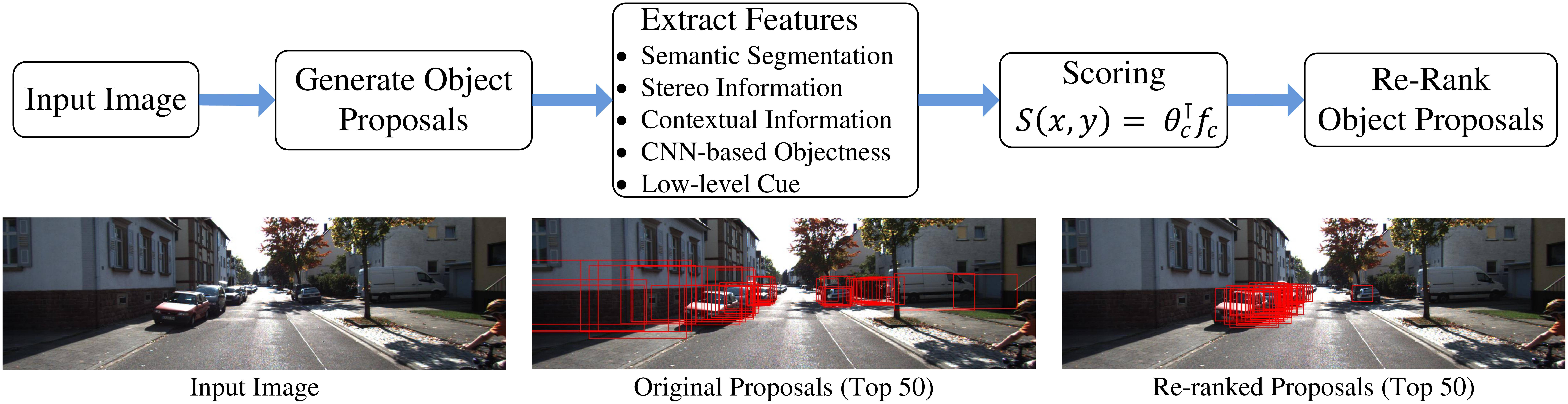}
 
    \end{overpic} 
   \caption{
   Overview of our re-ranking approach: Given an input image, we first use a generator to produce  a set of object proposals. 
   Then we extract effective features for each proposals, and score proposals by dot-product with learned weights. 
   Finally, We re-rank proposals using the computed scores. 
   Our approach select high quality proposals.}
\label{fig:overview}
\end{figure}

In this paper, we propose a class-specific object proposal re-ranking approach to score candidate proposals directly at the field of automatic driving. 
Figure \ref{fig:overview} shows the overview of our approach.
Given an input image and a set of object proposals, our approach contains the following three steps:

(1) Firstly, semantic segmentation, stereo information, contextual information, CNN-based objectness and low-level cue are extracted for each proposal. 
Specifically, we compute the semantic segmentation by
DeepLab \cite{DeepLab2015}, 
	in which the deep network is fine-tuned on the Cityscapes datasets \cite{cordts2015cityscapes}.
The disparity map is computed via the state-of-the-art CNN-based approach proposed by Zbontar et al. \cite{zbontar2015stereo},
	then we estimate the road plane with the computed disparity map. 
We use DeepBox  \cite{kuo2015deepbox} to compute
CNN-based objectness of each proposal.

(2) Secondly, Structured SVM \cite{tsochantaridis2004ssvm} is introduced to learn class-specific weights, 
	then we score each proposal by encoding extracted features.

(3) Finally, we re-rank object proposals depending on the computed scores.

Our experiments on KITTI show that our approach is able to significantly improve the recall performance of various object proposal methods. 
We achieve the best recall performance by merging with the 3DOP \cite{Chen20153DOP} method on \emph{Car}, \emph{Cyclist} and \emph{Pedestrian} categories. 
Furthermore, using 1,000 re-ranked 3DOP proposals per image obtains a slightly higher object detection average precision (AP) than using 5,000 3DOP proposals, 
	indicating that our approach selects more benificial proposals.

\section{Related Work}

   In recent years, object proposal has become very popular in object detection as an important pre-processing step. 
   Object proposal methods can be classified into three main categories: \emph{window scoring based methods}, \emph{grouping based methods}, and \emph{CNN-based methods}.

\textbf{Window scoring based methods:} Window scoring based methods attempt to score the objectness of each candidate proposal according to how likely it is to contain an object of interest. 
This category of methods first sample a set of candidate bounding boxes across scales and locations in an image, 
	and  measure the objectness scores based on scoring model and return top scoring candidates as proposals. 
Objectness \cite{alexe2012objectness} is one of the earliest proposal methods.
This method samples a set of proposals from salient locations in an image, 
	and then measures objectness of each proposals according to different low-level cues, such as saliency, colour, and edges.
BING \cite{cheng2014bing} proposes a real-time proposal generator by training a simple linear SVM on binary features, 
	the most obvious shortcoming of which is that it has a low localization accuracy.
EdgeBoxes \cite{zitnick2014edgeboxes} uses contour informations to score candidate windows without any parameter learning. 
In addition, it proposes a refinement process to promote localization. 
These methods are generally efficient, but suffer from poor localization quality.

\textbf{Grouping based methods:} Grouping based methods are segmentation-based approaches. 
They generally generate multiple hierarchical superpixels that are likely to contain objects for an image, 
	and then employ different grouping strategies to generate object proposals depending on different low-level cues, such as, colour, contour and texture. 
Selective Search \cite{van2011SS} greedily merges the most similar superpixels to generate proposals without learned parameters. 
This method has been widely applied by many state-of-the-art object detection methods. 
Multiscale Combinatorial Grouping (MCG) \cite{arbelaez2014MCG} generates  multi-scale hierarchical segmentations 
	and merges them based on the edge strength to obtain object proposals. 
Geodesic object proposals \cite{krahenbuhl2014geodesic} uses classifiers to place seeds for a geodesic distance transform 
	and selects object proposal by identifying critical level sets of the distance transforms.
Compared to window scoring based methods, 
	grouping based methods have better localization ability, but require more complex computation.

\textbf{CNN-based methods:} Benefit from the strong discrimination ability of Convolutional Neural Network (CNN), CNN-based methods directly generate high quality candidate proposals with a fully convolutional network (FCN).
Multibox \cite{erhan2014multiBox} trains a large CNN model to directly generate object proposals from images and ranks them depending on the predicted objectness scores. 
RPN \cite{ren2015fasterrcnn} uses a FCN to generate object proposals with a wide range of scales and aspect ratios. 
DeepBox \cite{kuo2015deepbox} uses a lightweight CNN to predict the objectness scores of candidate proposals and re-ranks them depending on the predicted objectness scores. 
These CNN-based methods achieve high recall with only a small number of proposals under loose criteria (e.g. IoU > 0.5),
	 but fails under strict criteria (e.g. IoU > 0.7).

\section{Re-ranking Object Proposals}
 In this section, we present a class-specific approach to re-rank object proposals for improving the recall rate. 
 Given a set of object proposals from an image, 
 	the goal of our re-ranking model is to select the proposals that are most likely to contain specific class of object. 
For each proposal, the score is assigned by encoding semantic segmentation,
	stereo information, contextual information, CNN-based objectness and low-level cue with class-specific weights. 
Then we re-rank proposals by sorting the computed scores. 
We use Structured SVM \cite{tsochantaridis2004ssvm} to learn class-specific weights for these features.

\subsection{Re-ranking Model}

 \begin{figure}
  \centering
  \subfigure{ \includegraphics[width=\linewidth]{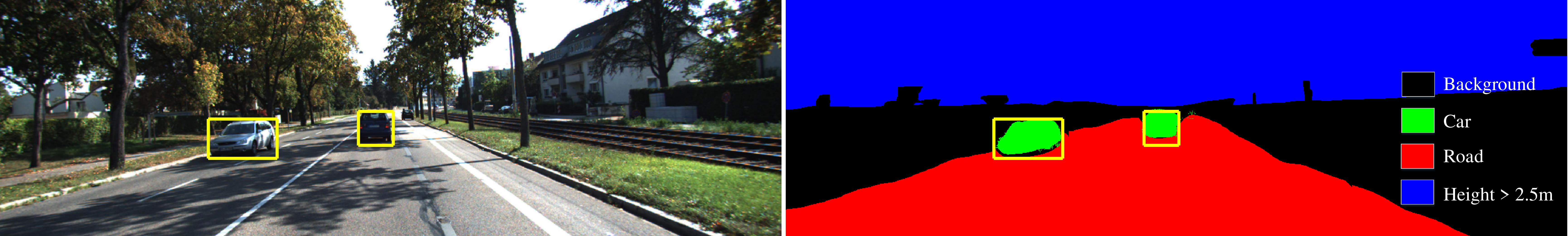}}  \vspace{-.2in}
  \caption{Example of object detection in the context of automatic driving. 
  \textbf{Left Image:} RGB image, \textbf{Right Image:} pixel-wise semantic segmentation and depth map. 
  Yellow rectangle denotes the high quality proposal.}
\label{fig:what-a-good-proposal}
\end{figure}

Figure \ref{fig:what-a-good-proposal} shows an example of object detection in the context of automatic driving. 
We observe that a high quality proposal (highly overlap with ground truth) has the following attributes:

(1) A high quality proposal is more likely to preserve certain class of object, that is, 
	it contains a larger proportion of such class of object than any other class inside the bounding box.

(2) A high quality proposal has a height restriction, 
	so that the height of objects inside the bounding box are lower than a constant threshold 
	(e.g. as shown in figure \ref{fig:what-a-good-proposal}, the height of a car is commonly not higher than 2 meters, 
	so the limited height of proposals should not be higher than a slightly larger constant, set to 2.5 meters, empirically).

(3) A high quality proposal partly contains road. 
	As objects are always on the road , a high quality proposal and the box under it commonly contain road component.

According to these attributes, 
	we formulate our scoring function by encoding semantic segmentation, stereo information, contextual information, CNN-based objectness, and low-level cue:
\begin{eqnarray}
 \begin{array}{l}
	\displaystyle S(\textbf{x}, \textbf{y}) = \quad \theta^\top_{c,sem}f_{c,sem}(\textbf{x},\textbf{y}) + \theta^\top_{c,hei}f_{c,hei}(\textbf{x},\textbf{y}) + \\ \\
	\displaystyle \quad \qquad \qquad  \theta^\top_{c,cont}f_{c,cont}(\textbf{x},\textbf{y}) 
	+ \theta^\top_{c,cnn}f_{c,cnn}(\textbf{x},\textbf{y}) +  \theta^\top_{c,low}f_{c,low}(\textbf{x},\textbf{y}) 
	
 \end{array}
	\label{eq:score}
\end{eqnarray}
Where,  $\textbf{x}$ denotes input image, and $\textbf{y}$ is a proposal in the set of proposals, $\textbf{Y} = \{ \textbf{y}^{1},...,\textbf{y}^{n}\}$, $n$ is the number of proposals. 
Note that, our score depends on the object class via class-specific weights $\theta_c = \left \{ \theta_{c,sem}, \theta_{c,hei}, \theta_{c,cont}, \theta_{c,cnn}, \theta_{c,low} \right \} $, 
	which are learned using structured SVM  \cite{tsochantaridis2004ssvm}. 
We next describe details of each feature.

\subsection{Re-ranking Features}
 
\textbf{Semantic Segmentation:} 
Taking advantage of pixel-wise semantic segmentation,
	 this feature is two dimensional, of which the first dimension is to encourage the existence of an object inside the box and the second to ensure that of the road.
 The first dimension counts the ratio of pixels labeled as the specific class: 
\begin{eqnarray*}
   \begin{array}{l}
	f_{c,seg,c}(\textbf{x},\textbf{y}) = \frac{\sum_{i\in \Omega (\textbf{y})}Seg_c(i)}{\left | \Omega (\textbf{y}) \right |}
	\end{array}
	\label{eq:seg}
\end{eqnarray*}
 where $\Omega (\textbf{y})$ is the set of pixels in the bounding box $\textbf{y}$, and $Seg_c(i)$ denotes the segmentation mask for class $c$.
The other feature computes the ratio of pixels labeled as the road:
\begin{eqnarray*}
 \begin{array}{l}
 
	f_{c,seg,road}(\textbf{y}) = \frac{\sum_{i\in \Omega (\textbf{y})}Seg_{road}(i)}{\left | \Omega (\textbf{y}) \right |}
 \end{array}
	\label{eq:road}
\end{eqnarray*}
where $Seg_{road}$ denotes the segmentation mask for road.
Note that this feature can be computed very efficiently by using as many integral images as classes.
We use DeepLab \cite{DeepLab2015} to compute pixel-wise semantic segmentation. 
Deeplab is a semantic segmentation model that uses convolutional neural networks and fully-connected conditional random fields to produce accurate segmentation maps. 
Since very few semantic annotations are available for KITTI, 
	we train the Deeplab model on the Cityscapes \cite{cordts2015cityscapes} dataset.
The Cityscapes dataset is similar to the KITTI dataset which contains dense pixel annotations of 19 semantic classes such as road, car, pedestrian, etc.

\textbf{Height:} This feature encodes the fact that the height of the pixels in the bounding box should not be higher than the height of the object class $c$. 
To minimize the presence of excessively high pixels inside the bounding box, 
	we get this feature based on the percentage of pixels for which the height exceed a threshold $\tau $. 
\begin{eqnarray*}
 \begin{array}{l}
	f_{c,hei}(\textbf{y}) = - \frac{\sum_{i\in \Omega (\textbf{y})}H(i)}{\left | \Omega (\textbf{y}) \right |}	
	\end{array}
	\label{eq:height}
\end{eqnarray*}
where  $H(i)$ is an indicator, with $H(i) = 1$ if the height of $i$ is larger than a threshold $\tau$, $H(i) = 0$ otherwise, in this paper, we set $\tau = 2.5$ $m$. 
This feature is inversely proportional to $S(\textbf{x},\textbf{y})$. 
We assume a stereo image pair as input and compute depth map via the state-of-the-art approach proposed in \cite{zbontar2015stereo}, 
	and then obtain the height for each pixel with the computed depth map. 
This feature can be very efficiently computed using integral images. 

\textbf{Context:} This feature encodes the contextual road information and the contextual height information. 
In the context of automatic driving, cars and pedestrians are on the road, 
	so we can see road below them, as well as the height below them would not exceed the object. 
We use a rectangle below the bounding box as the contextual region.
We set its height as one-third the height of the box, and use the same width. 
We then compute semantic segmentation feature and height feature of the contextual region. 
Note that, we only compute the second dimension of semantic segmentation feature, 
	i.e., we only ensure the presence of road in the contextual region.

\textbf{CNN-based Objectness:} 
We use DeepBox \cite{kuo2015deepbox} to compute the CNN-based objectness of proposals. DeepBox is a lightweight CNN model uses a novel four-layer CNN architecture to compute the objectness score of object proposals. We pre-train the DeepBox model on  PASCAL VOC \cite{everingham2015pascal} + COCO \cite{lin2014miCOCO}.
This feature can efficiently prune away easily distinguished false positives, 
	enabling our model to focus on proposals that are more likely to contain objects.

\textbf{Low-level Cue:} This feature is the ranking score derived from the object proposal generator that produce candidate proposals $\textbf{y}$. 
Given that, some object proposal generators do not produce ranking scores for proposals, 
	such as selective search, we give each proposal an identical low-level score for these generators.

\subsection{Re-ranking Loss}
 In order to train the weights, we define the task loss function $\Delta (\textbf{y}_{gt},\textbf{y})$ as the Intersection-over-Union (IoU) between the set of GT boxes, $\textbf{y}_{gt}$, and  candidate proposals $\textbf{y}$:
\begin{eqnarray*}
 \begin{array}{l}
	\Delta (\textbf{y}_{gt},\textbf{y}) = \frac{area(\textbf{y}_{gt} \cap \textbf{y} )}{area(\textbf{y}_{gt}\cup \textbf{y}}
	
	\end{array}
	\label{eq:loss}
\end{eqnarray*}
 where $area(\textbf{y}_{gt}\cap \textbf{y} )$ denotes the intersection of the  ground truth and candidate proposal
 bounding boxes, and $area(\textbf{y}_{gt} \cup \textbf{y} )$ their union.

\subsection{Parameter learning}
We learn the weights $\theta_c = \left \{ \theta_{c,sem}, \theta_{c,hei}, \theta_{c,cont}, \theta_{c,cnn}, \theta_{c,low} \right \} $ of the scoring model by solving the following Structured SVM \cite{tsochantaridis2004ssvm} Quadratic Program:
\begin{eqnarray}
 \centering
 \begin{array}{l} 
	\displaystyle \rm \underset{\theta,\xi_i}{\min} \quad \left \| \theta  \right \|_2^2
	+ C \sum_{i =1 }^{N} \xi_i \\ \\
	 \text{s.t.} \quad  \theta^\top \left ( f(\textbf{x}^{i}, \textbf{y}_{gt}) - f(\textbf{x}^{i}, \textbf{y}^i) \right )  \geq  1- \frac{\xi_i}{ \Delta (\textbf{y}_{gt},\textbf{y})}, \xi_i \geq 0, \forall \textbf{y}_{gt} \setminus  \textbf{y}^i 
	
 \end{array}
	\label{eq:ssvm}
\end{eqnarray}
We solve \ref{eq:ssvm} via the parallel cutting plane of \cite{schwing2013parallelcutting}. 
At testing time, the re-ranking process is simple and efficient.
We first compute the  features of each object proposal, 
	and then the score is computed by applying dot-product between the features and  the learned weights, 
	and finally, the re-ranked proposals are generated by sorting according to the computed scores.

\begin{figure}[!t]
  \centering
  \subfigure[Car-Easy]{ \includegraphics[width=.32\linewidth]{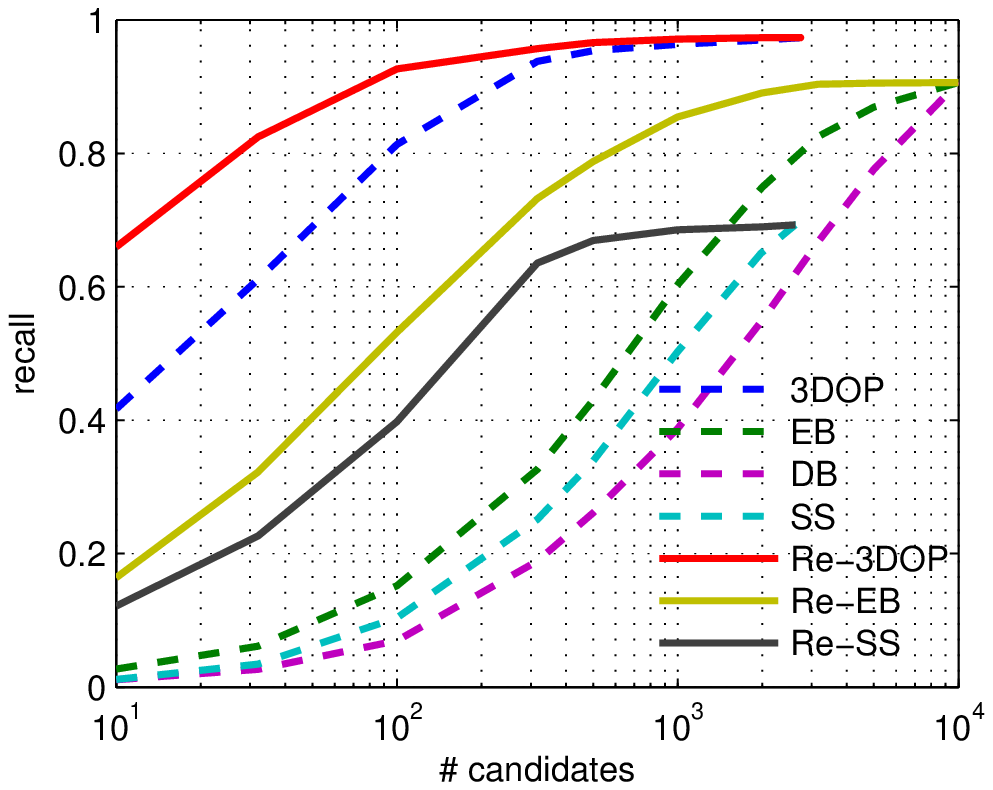}}\vspace{-.05in}
  \subfigure[Car-Moderate]{ \includegraphics[width=.32\linewidth]{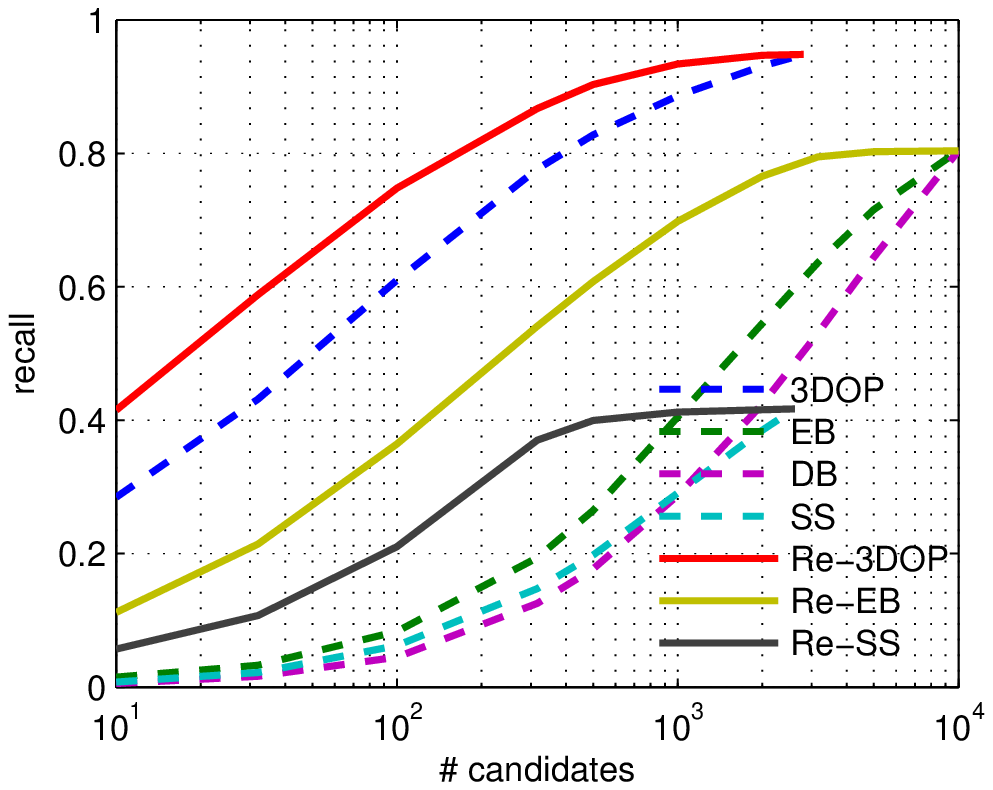}}\vspace{-.05in}
  \subfigure[Car-Hard]{ \includegraphics[width=.32\linewidth]{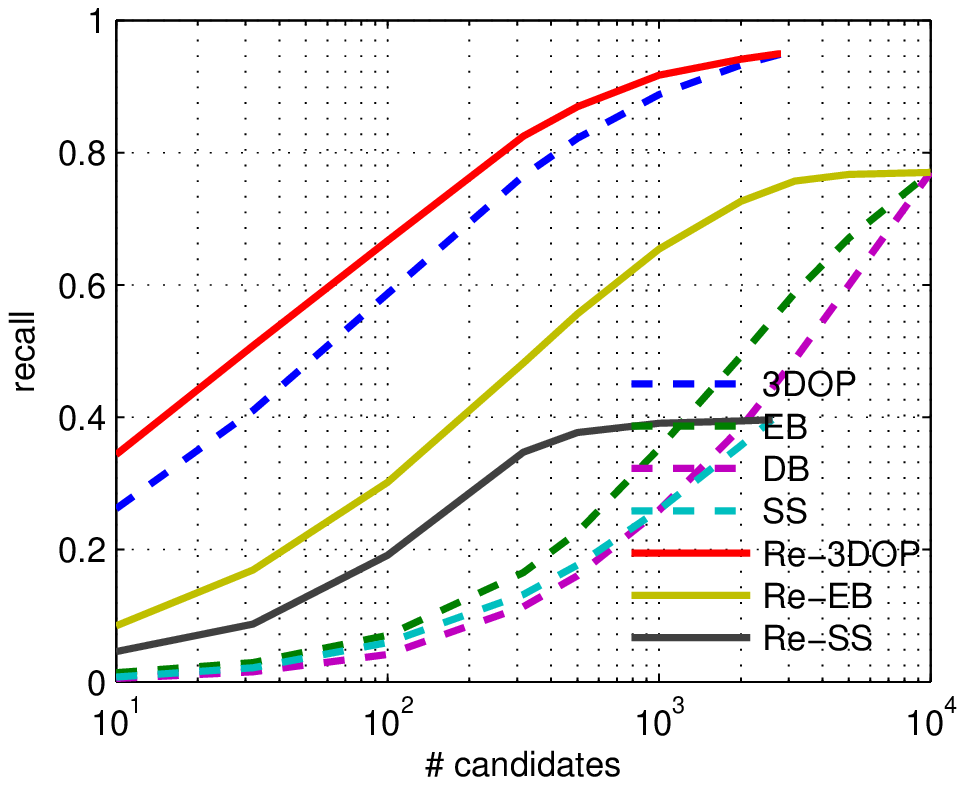}}\vspace{-.05in}

     \subfigure[Pedestrian-Easy]{ \includegraphics[width=.32\linewidth]{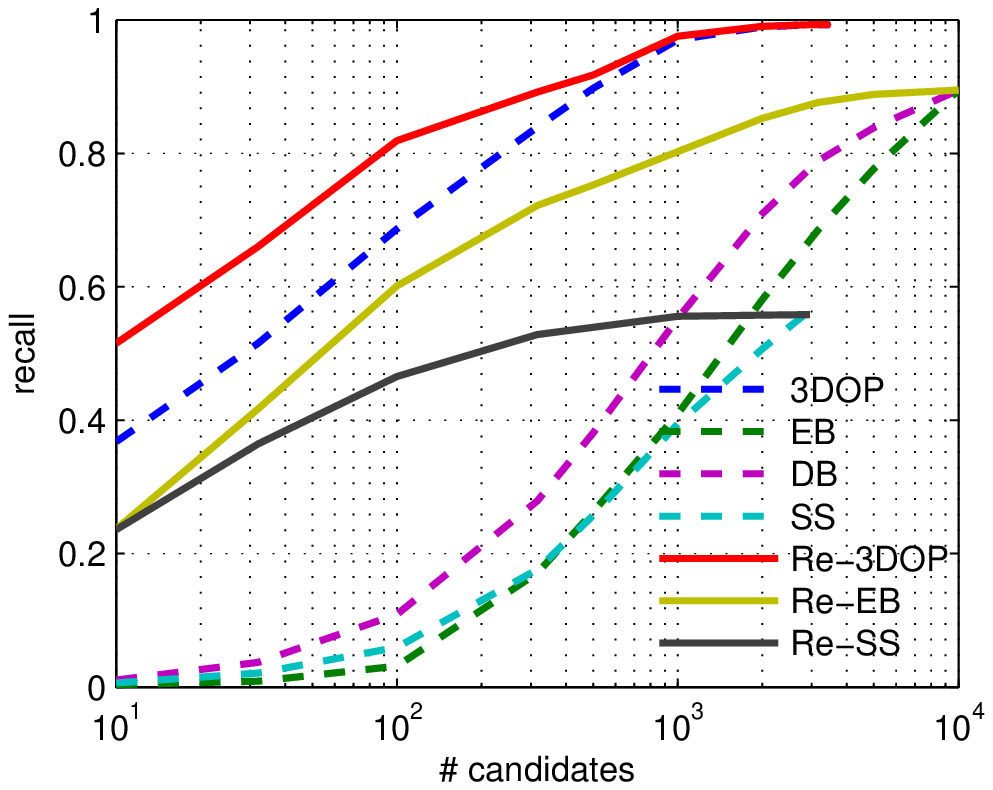}}\vspace{-.05in}
  \subfigure[Pedestrian-Moderate]{ \includegraphics[width=.32\linewidth]{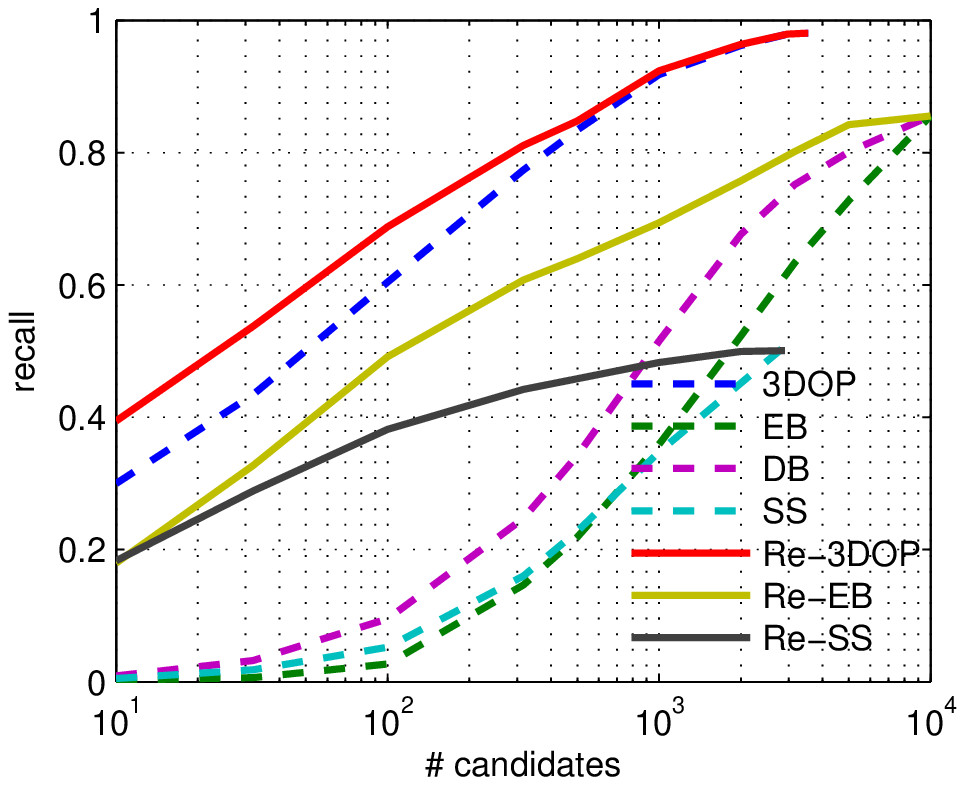}}\vspace{-.05in}
  \subfigure[Pedestrian-Hard]{ \includegraphics[width=.32\linewidth]{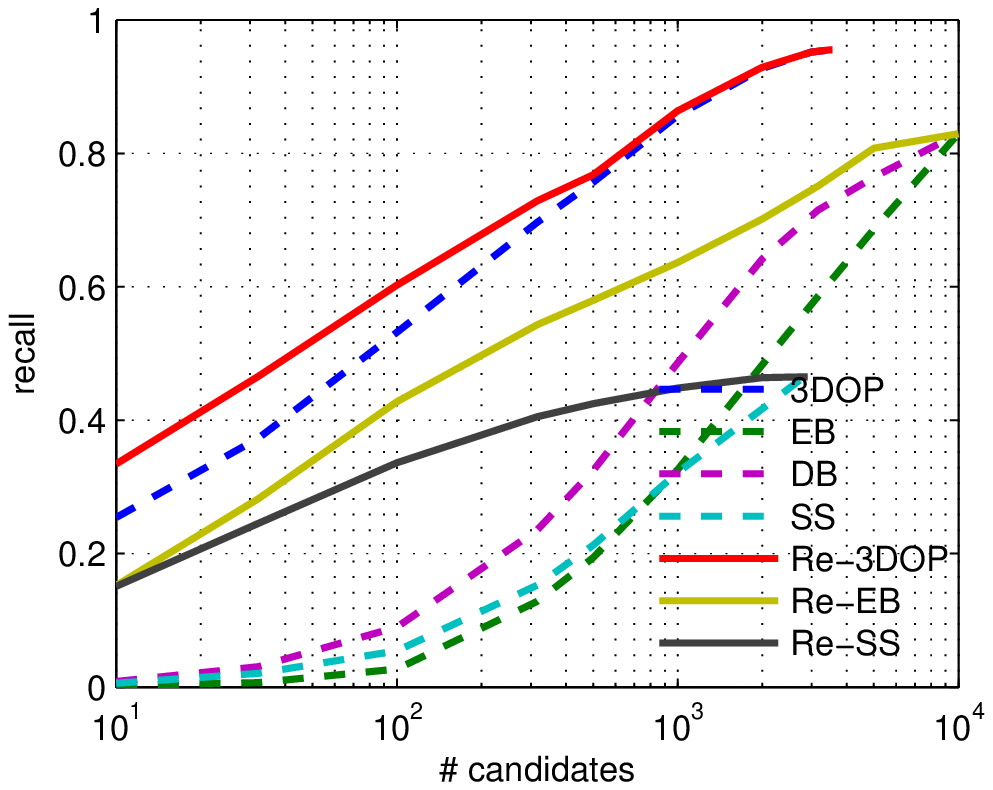}}\vspace{-.05in}

   \subfigure[Cyclist-Easy]{ \includegraphics[width=.32\linewidth]{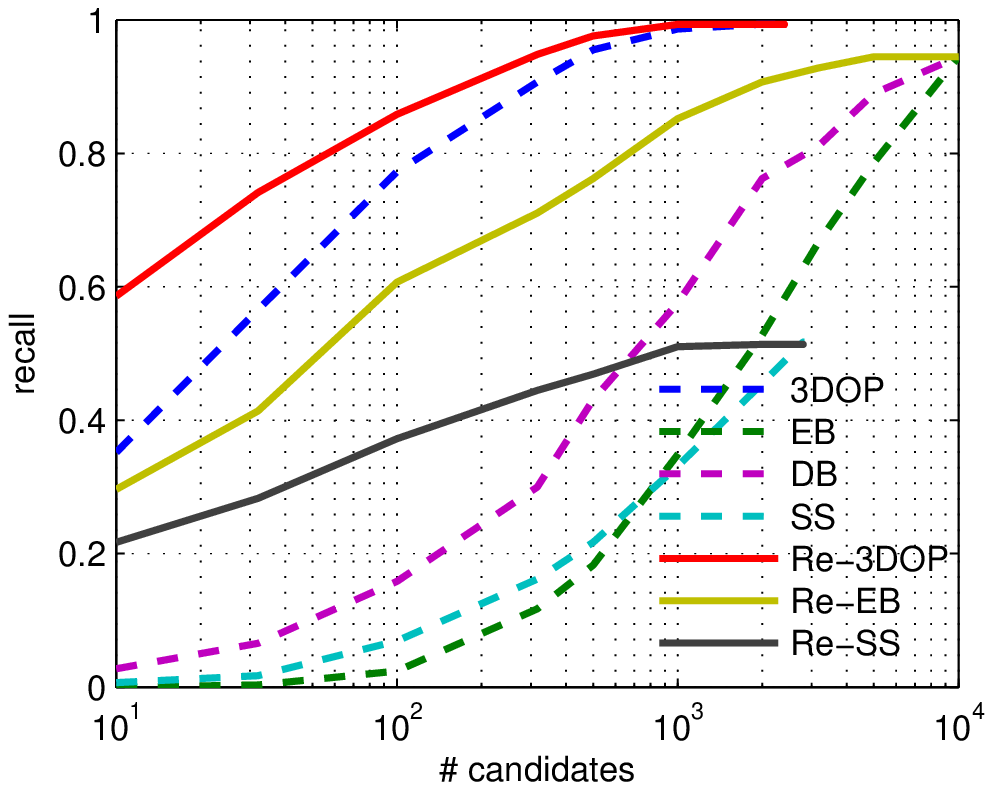}}\vspace{-.05in}
  \subfigure[Cyclist-Moderate]{ \includegraphics[width=.32\linewidth]{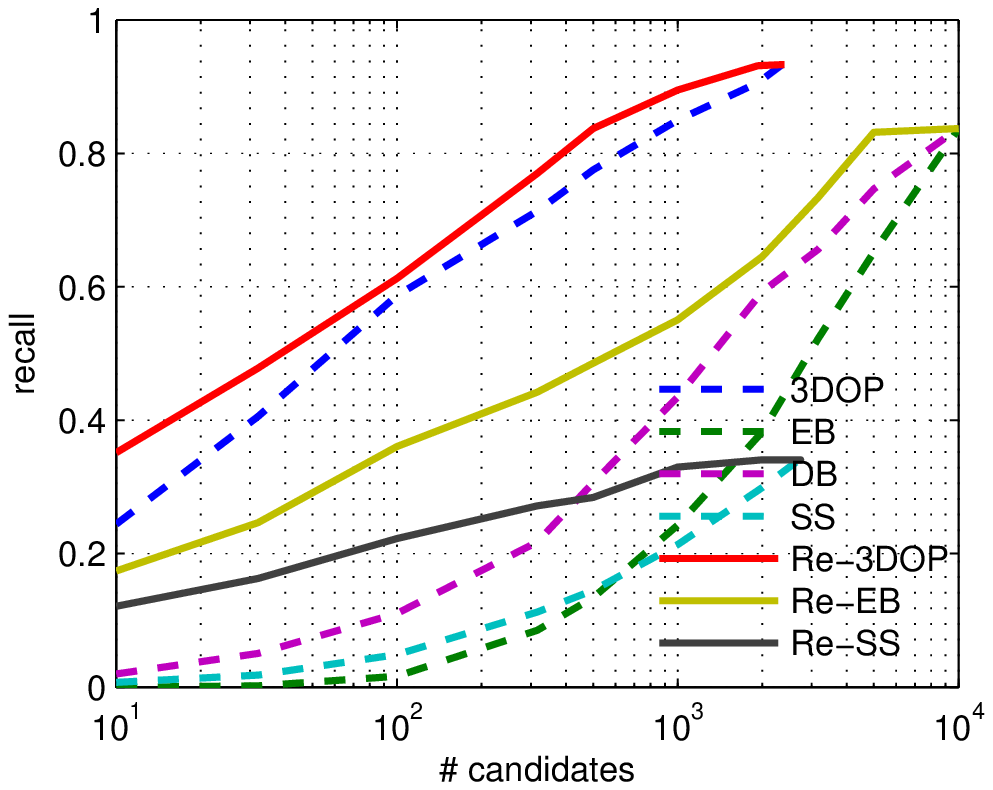}}\vspace{-.05in}
  \subfigure[Cyclist-Hard]{ \includegraphics[width=.32\linewidth]{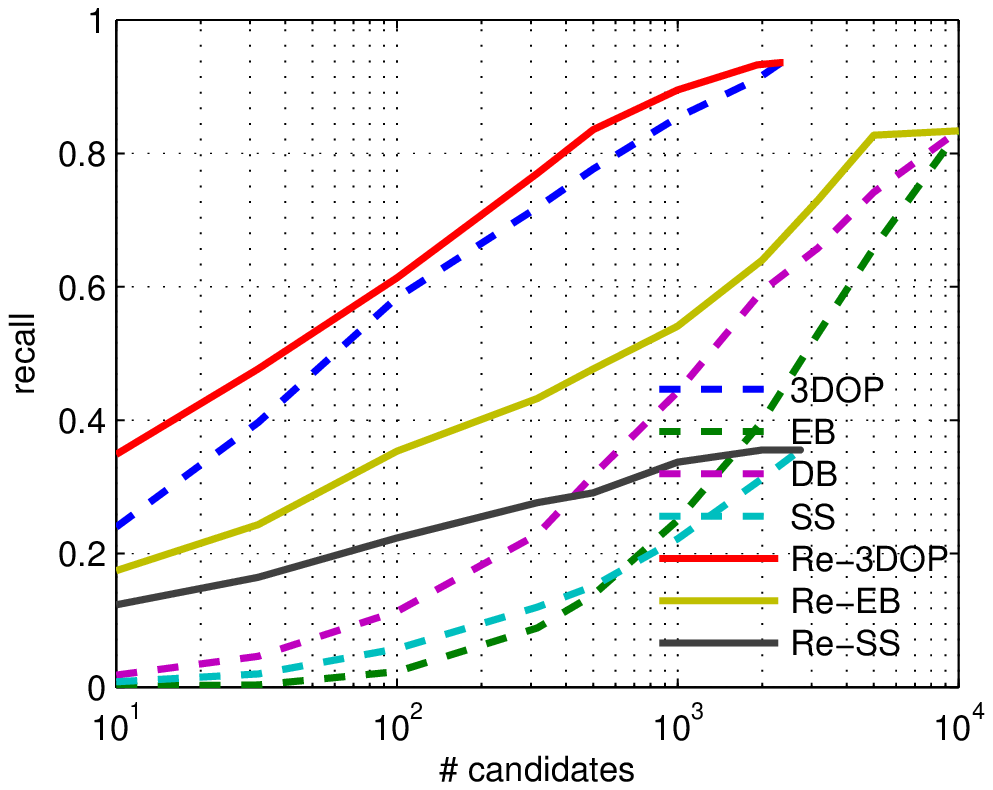}}\vspace{-.05in}
     
  \caption{\textbf{Recall VS Candidates:} 
  We use an IoU threshold of $0.7$ for Car, and 0.5 for Pedestrian and Cyclist. 
  The baseline methods are drawn in dashed lines.
  }\label{fig:recall-proposal}
\end{figure}

  \begin{figure}[!htb]
  \centering
  \subfigure[Car-Easy]{ \includegraphics[width=.32\linewidth]{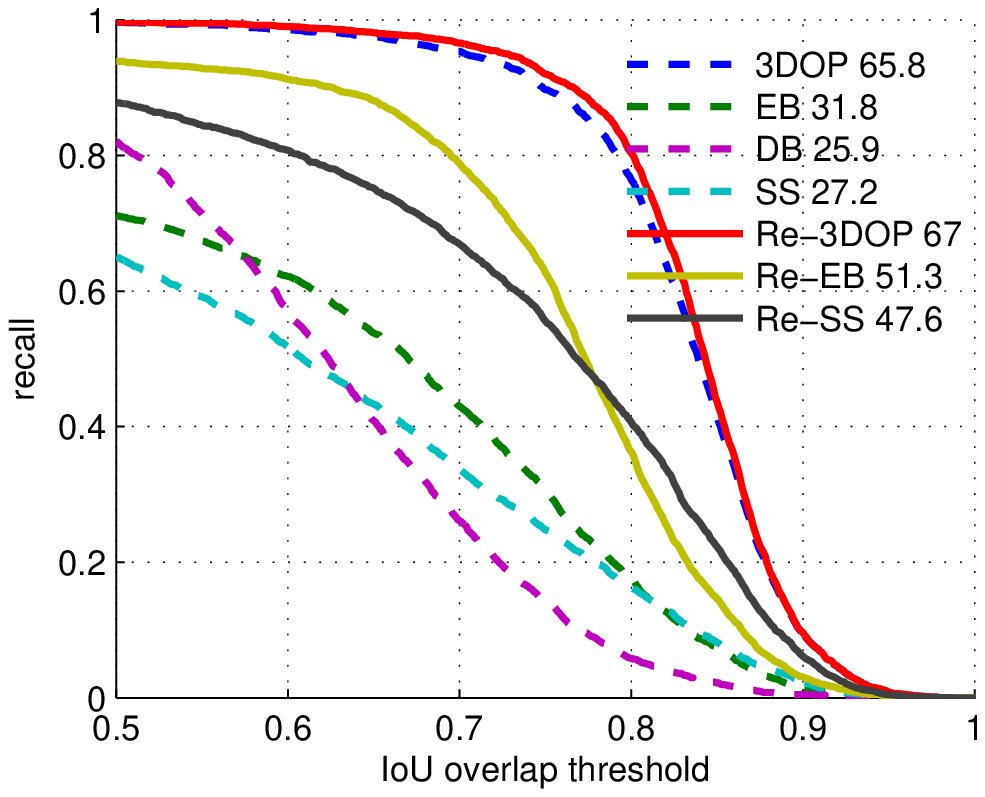}}\vspace{-.05in}
  \subfigure[Car-Moderate]{ \includegraphics[width=.32\linewidth]{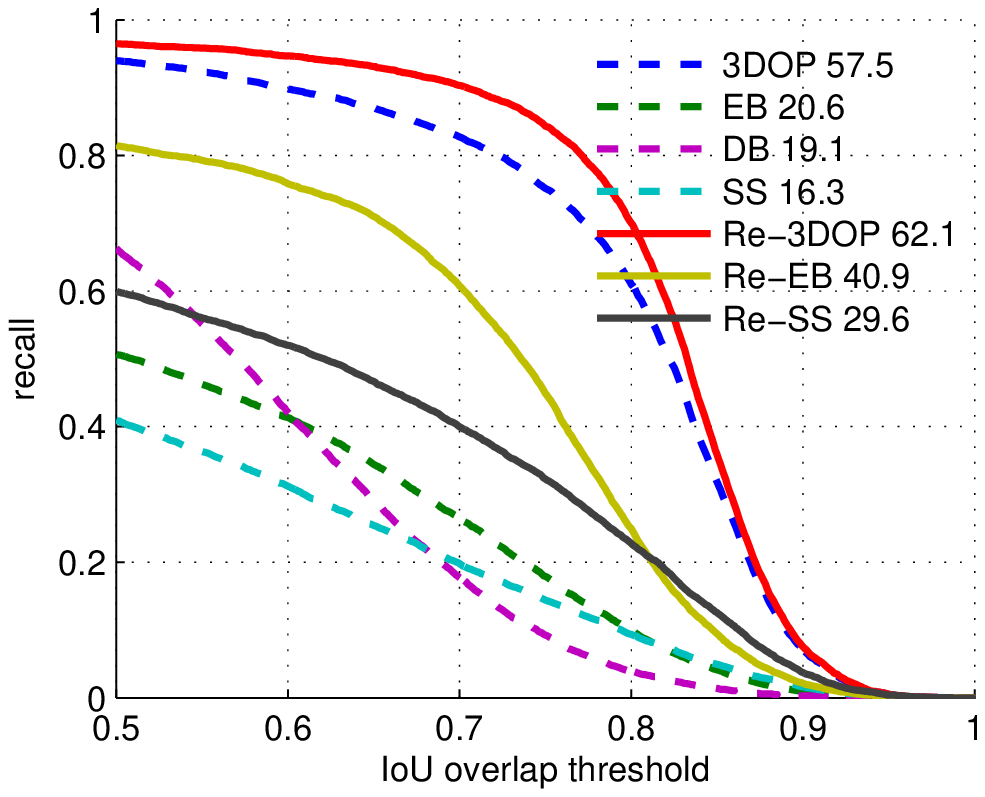}}\vspace{-.05in}
  \subfigure[Car-Hard]{ \includegraphics[width=.32\linewidth]{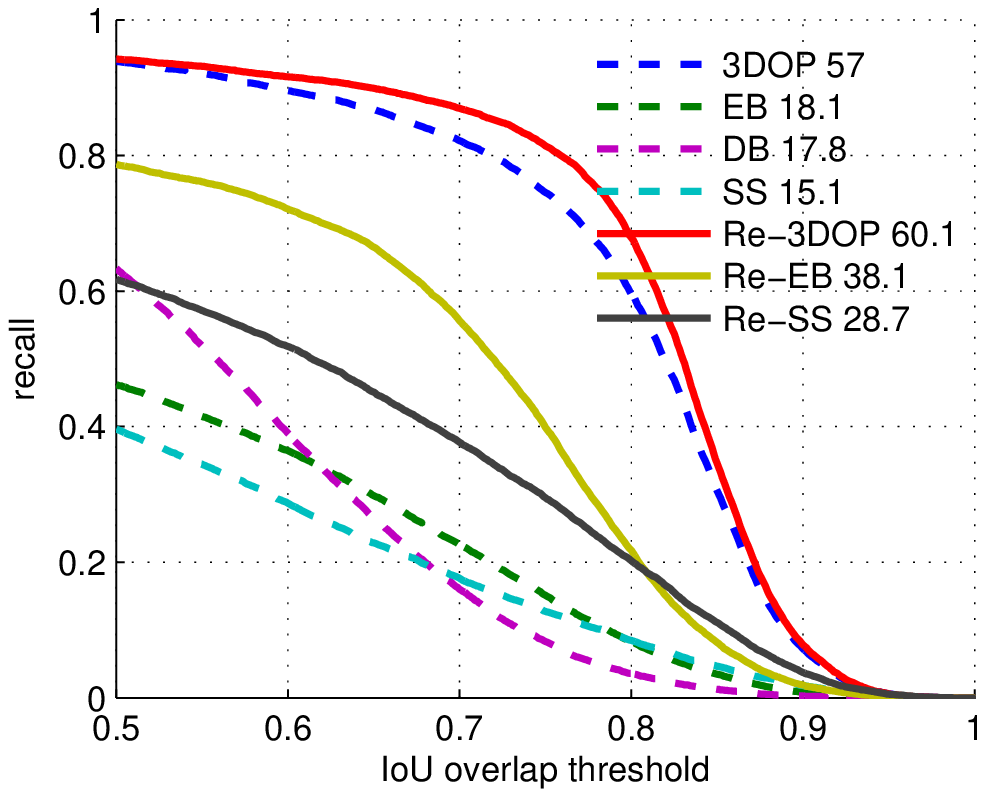}}\vspace{-.05in}

     \subfigure[Pedestrian-Easy]{ \includegraphics[width=.32\linewidth]{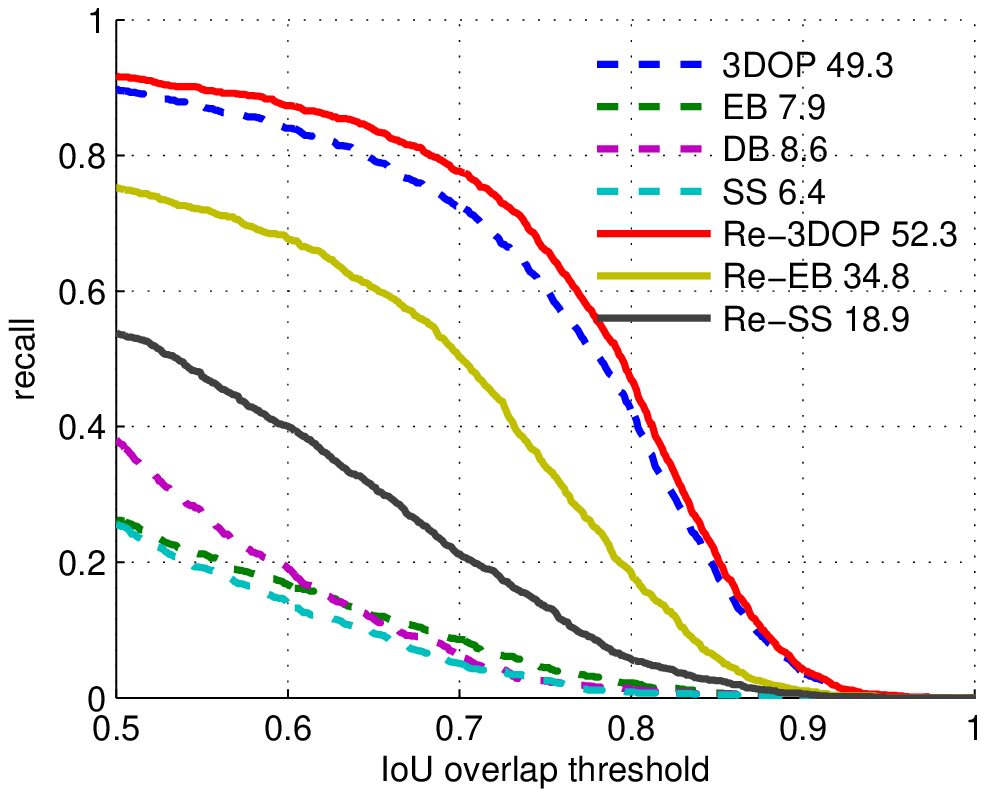}}\vspace{-.05in}
  \subfigure[Pedestrian-Moderate]{ \includegraphics[width=.32\linewidth]{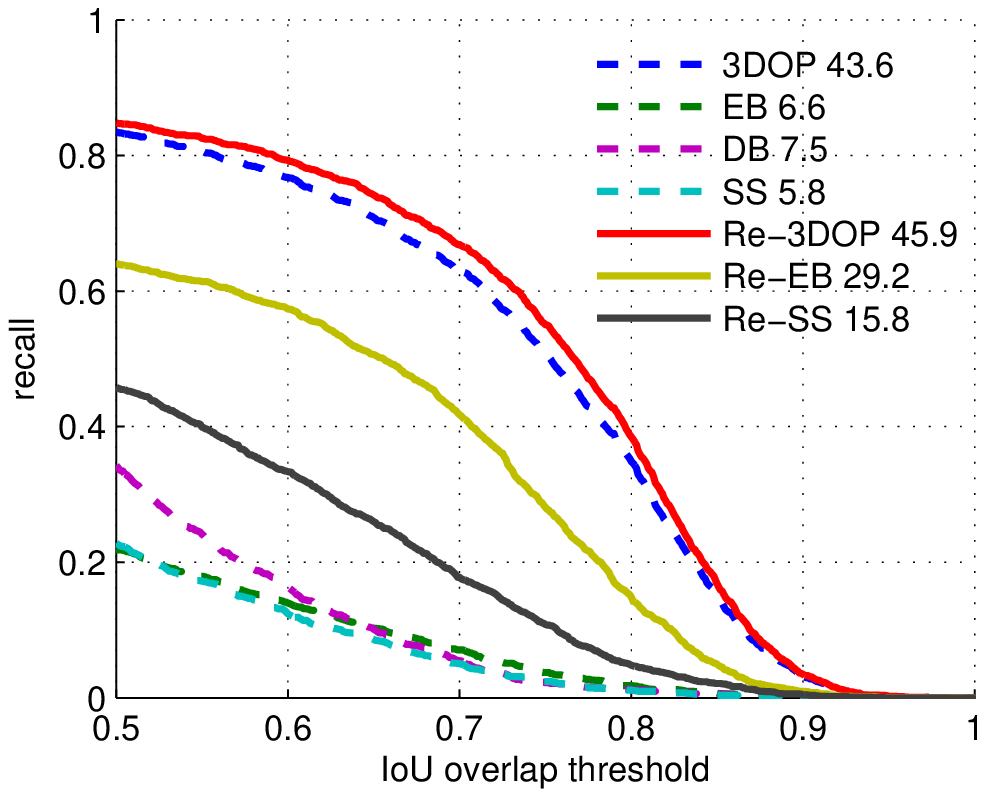}}\vspace{-.05in}
  \subfigure[Pedestrian-Hard]{ \includegraphics[width=.32\linewidth]{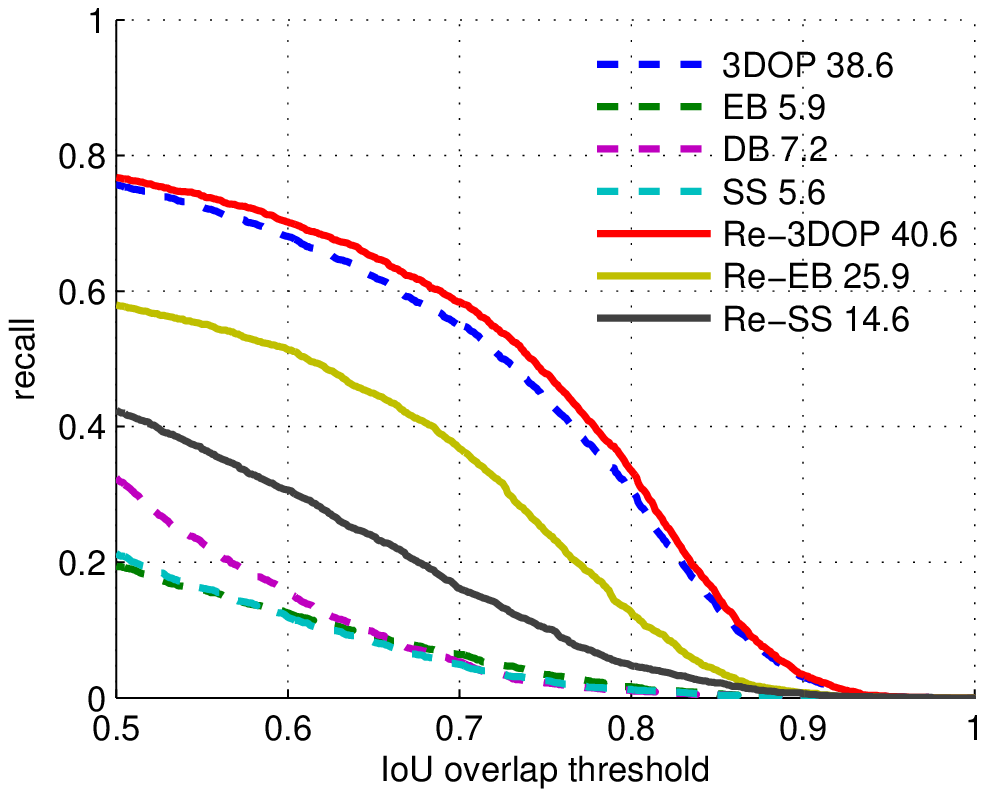}}\vspace{-.05in}

   \subfigure[Cyclist-Easy]{ \includegraphics[width=.32\linewidth]{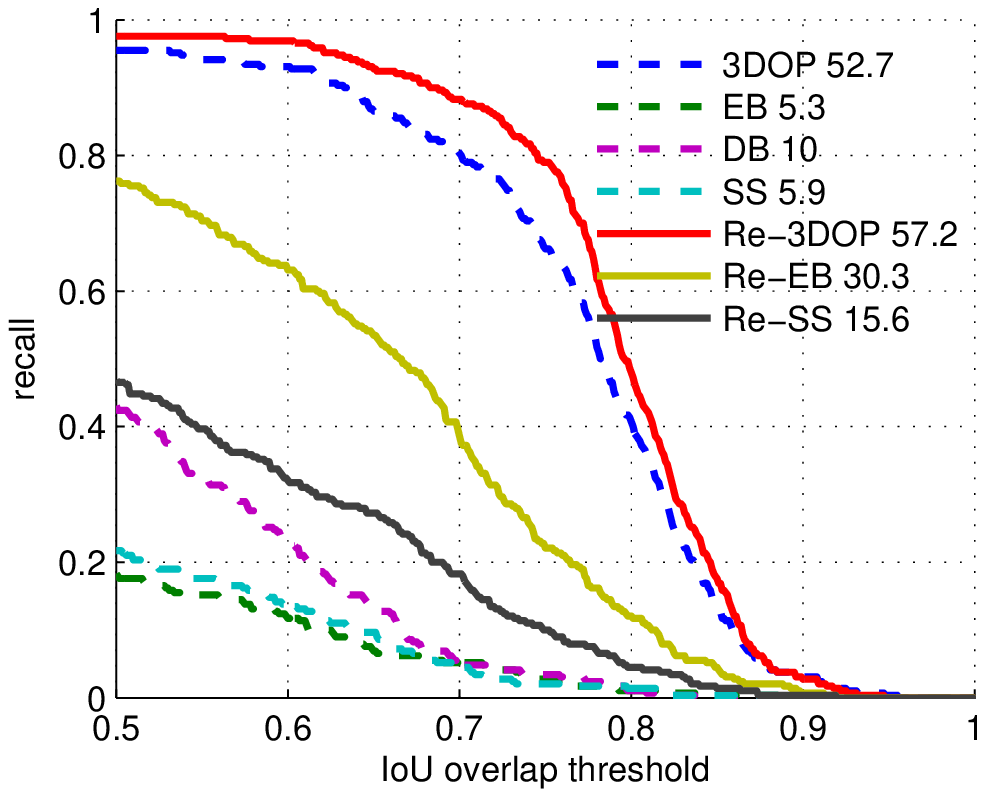}}\vspace{-.03in}
  \subfigure[Cyclist-Moderate]{ \includegraphics[width=.32\linewidth]{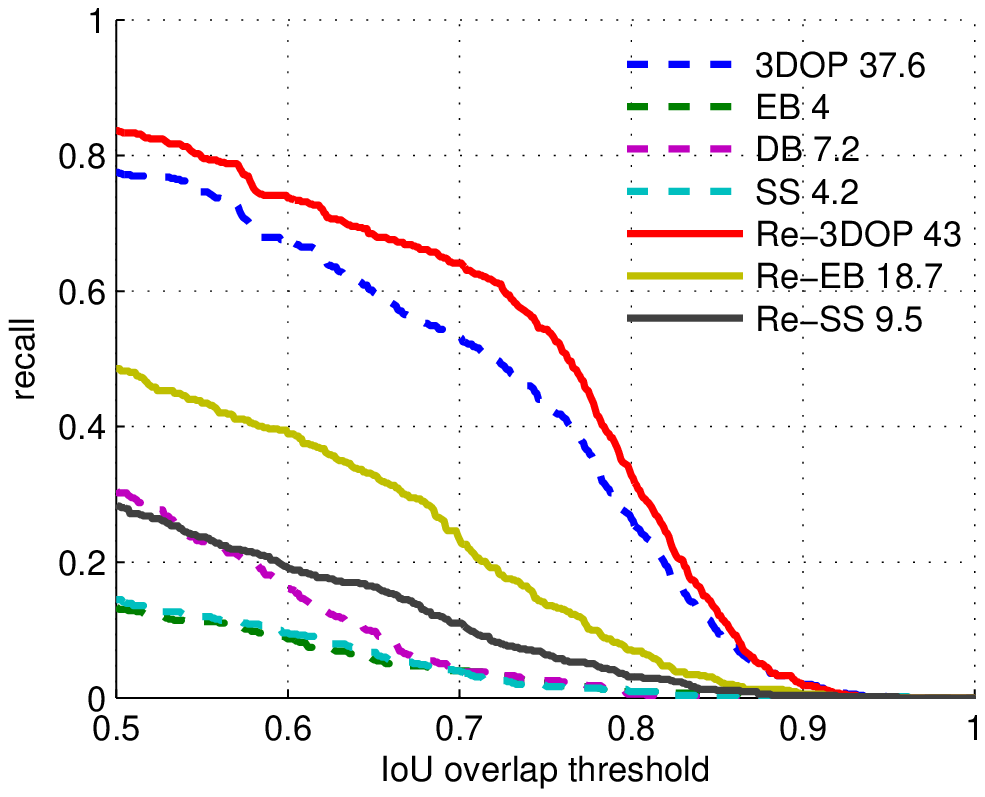}}\vspace{-.03in}
  \subfigure[Cyclist-Hard]{ \includegraphics[width=.32\linewidth]{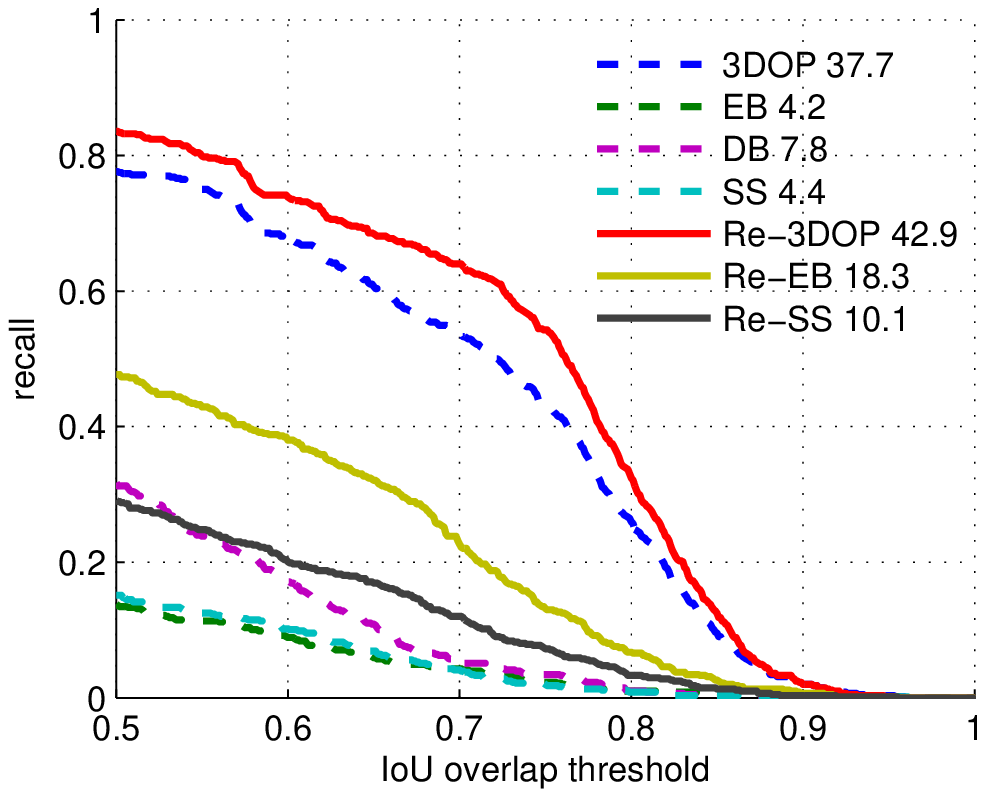}}\vspace{-.03in}

  \caption{\textbf{Recall vs IoU at 500 proposals.} 
Our approach successfully improves recall rate across various IoU threshold, especially in a strict criteria.
  }\label{fig:recall_500}\vspace{-.2in}
\end{figure}

  \begin{figure}[t]
  \centering
  \subfigure[Car]{ \includegraphics[width=.32\linewidth]{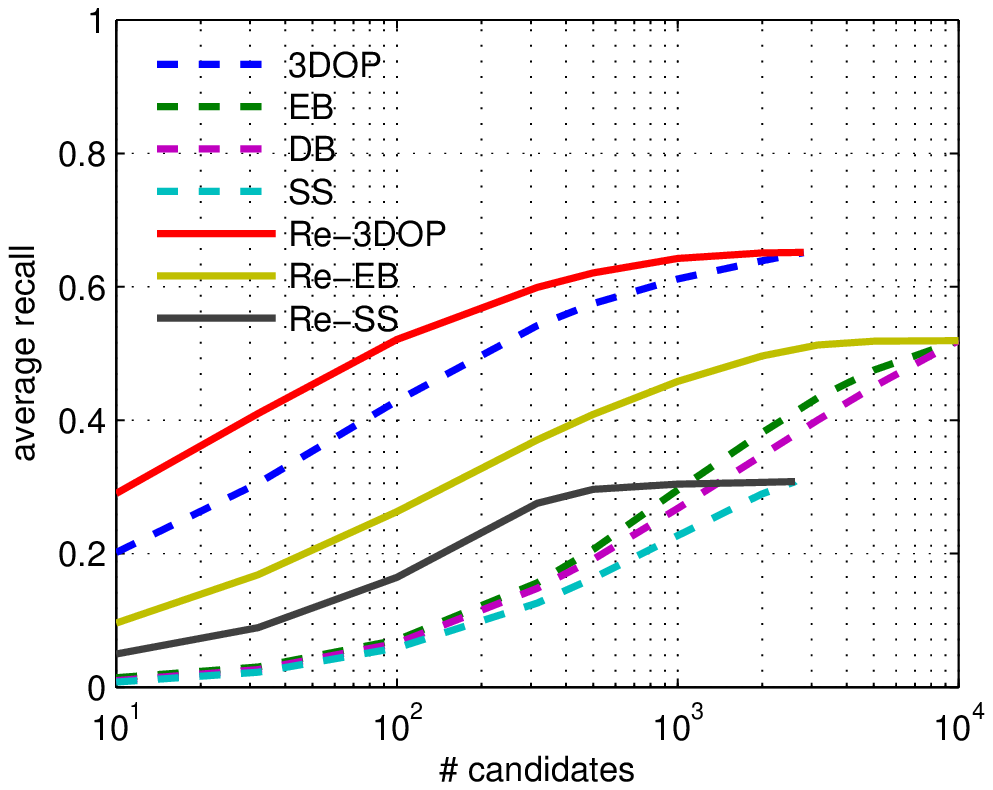}}\vspace{-.05in}
  \subfigure[Pedestrian]{ \includegraphics[width=.32\linewidth]{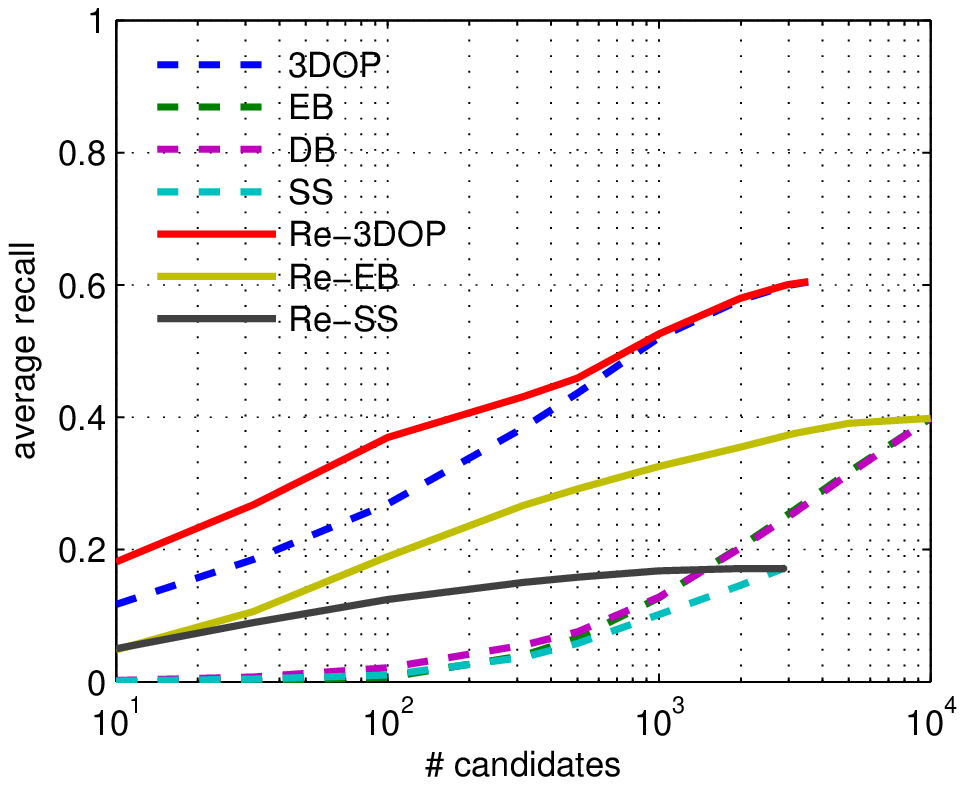}}\vspace{-.05in}
  \subfigure[Cyclist]{ \includegraphics[width=.32\linewidth]{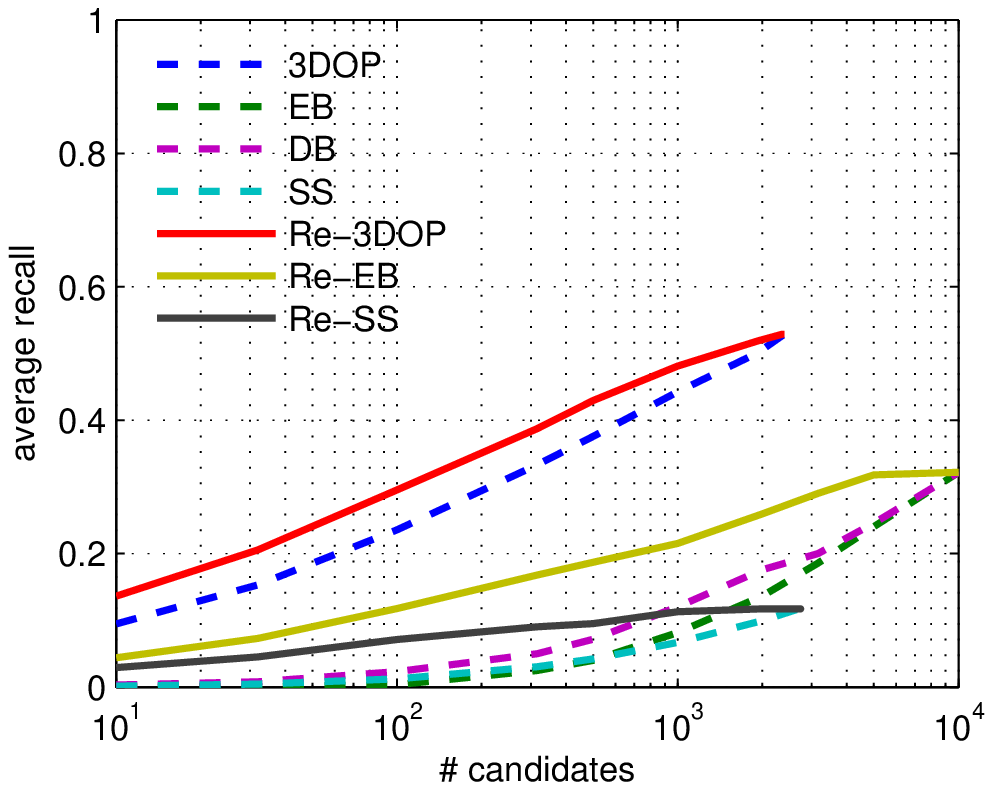}}
  \caption{\textbf{Average Recall (AR) vs Number of Candidates on moderate difficulty level.}
  Our approach consistently improves the AR, especially at a small number of proposals.
  }\label{fig:ar-proposal}
\end{figure}

\begin{table}
\footnotesize
\begin{center}
\begin{tabular}{c|c|c|c|c|c|c|c|c|c|c}
\hline
\hline
\multirow{2}{*}{Proposals}  & \multicolumn{10}{c}{\# candidates}  \\
\cline{2-11}
& 10 & 50 & 100  & 500 & 1K & 1.5K & 2K & 3K & 4K & 5K \\
\hline

3DOP \cite{Chen20153DOP} & 35.73 & 53.74 & 70.26  & 80.01 & 87.30 & 87.80 & 88.07 & 88.25 & \textbf{88.26} & \textbf{88.26} \\

Re-3DOP & 53.33 & 71.36 & 79.75  & 87.89 & 88.34 & 88.34 & \textbf{88.37}  & 88.35 & 88.35 & 88.35 \\

SS \cite{van2011SS} & 9.09 & 9.09 & 17.25  & 34.34 & 43.00 & 44.28 & \textbf{51.43}  & - & - & - \\

Re-SS & 16.34 & 25.43 & 33.80  & 51.51 & 51.86 & 51.87 & \textbf{51.89}  & - & - & - \\

EB \cite{zitnick2014edgeboxes} & 9.09 & 15.61 & 17.79  & 43.25 & 58.30 & 60.94 & 67.64 & 73.86 & 75.73 & \textbf{76.62} \\

DB \cite{kuo2015deepbox} & 9.09 & 14.30 & 21.50  & 40.14 & 55.04 & 62.63 & 65.40 & 71.34 & 73.78 & \textbf{75.40} \\

Re-EB & 25.13 & 42.34 & 51.08 & 68.44 & 74.84 & 76.81 & 77.12 & 77.60 & \textbf{81.13} & 81.03 \\
\hline
\hline

\end{tabular}
\end{center}
\caption{Average Precision (AP) (in \%) of object detection with different number of candidate proposals for \emph{Car} on moderate difficulty level on the validation set of KITTI Object Detection dataset.}
\label{tabel:dtection AP}
\end{table}

\begin{table}[ht]
\footnotesize
\begin{center}
\begin{tabular}{c|c|c|c|c|c}
\hline
\hline

 &\multicolumn{4}{c|}{Process} & \multirow{2}{*}{Totals}  \\
 \cline{2-5}
 
 & Semantic Segmentation & Depth Maps & CNN-based Objectness & Others\\
\hline

Times (s) &  0.21  & 0.35  & 0.2  & 0.03  & 0.79  \\
\hline
\hline
\end{tabular}
\end{center}
\caption{Running time of each step of our approach. }
\label{tabel:run times}
\end{table}

\begin{figure}[t]
  \centering
  
  \subfigure{ \includegraphics[width=\linewidth]{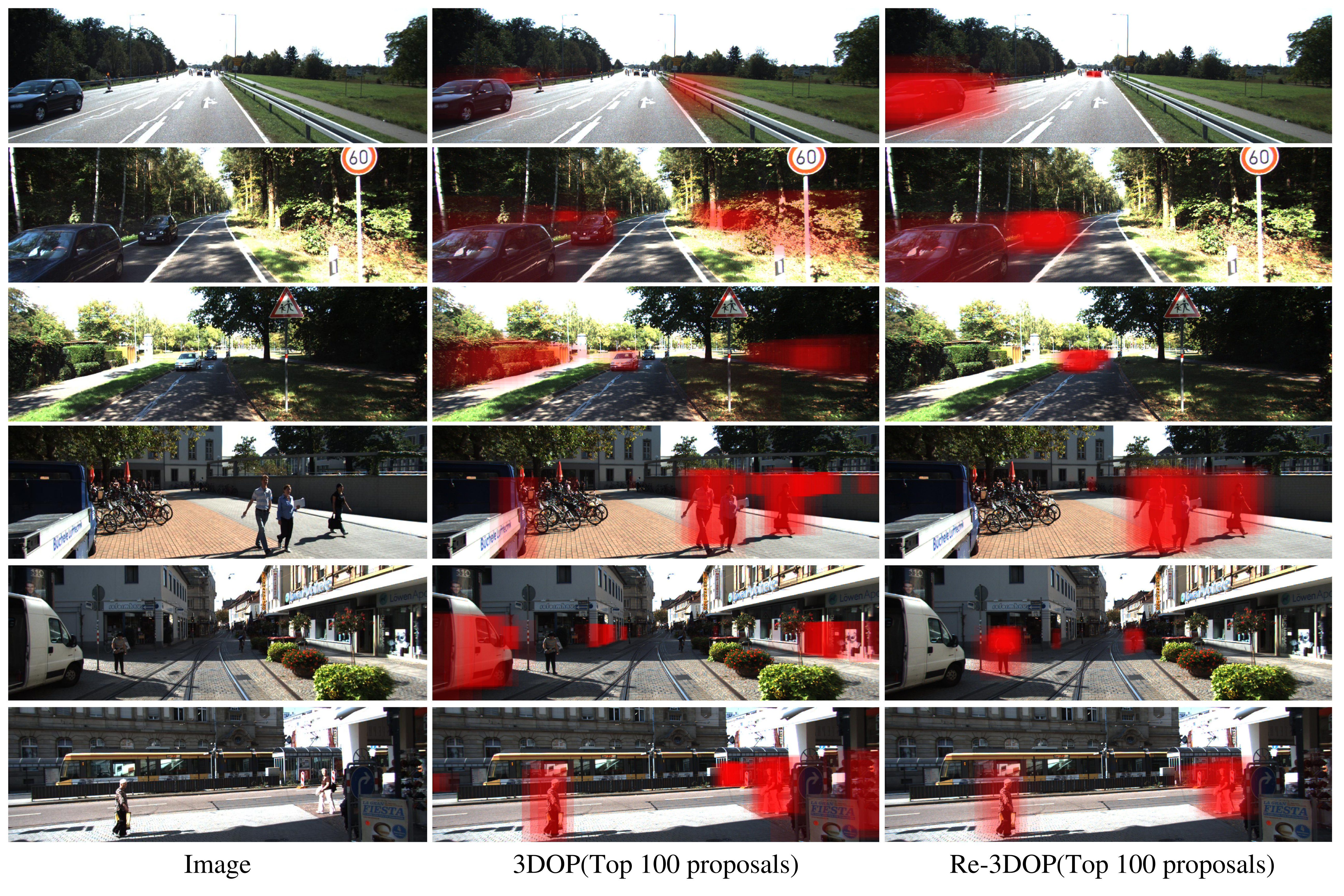}}
  
  \caption{Examples of the distribution of the top 100 scoring proposals shown by pasting a red box for each proposal.
From left to right: Images, top 100 proposals of 3DOP, top 100 proposals of Re-3DOP.}
\label{fig:exapmle}
\end{figure}

\section{Experiments}
\textbf{Dataset:} 
We evaluate our approach on the KITTI detection benchmark dataset \cite{geiger2012KITTI}. 
The KITTI estimation dataset consists of three categories: \emph{Car}, \emph{Pedestrian}, and \emph{Cyclist}, 
	with 7,481 images for training  and 7,518 images for testing, 
	and a total of 80,256 labeled objects. 
Evaluation for each class has three difficulty levels: Easy, Moderate, and Hard, 
	which are defined in term of the occlusion, size and truncation levels of objects. 
Since the ground truth labels of  the test set are not publicly available for researchers, following \cite{Chen20153DOP}, 
	we partition the KITTI training images set into training and validation sets to evaluate our approach, 
	which consist of 3,712 images and 3,769 images respectively. 
We insure that images from the same video are not simultaneously present in the training and validation sets.
We use the training set to learn the parameters by using structured SVM \cite{tsochantaridis2004ssvm}, 
	and evaluate the recall performance of proposals on the validation set.

\textbf{Evaluation Metrics:}
To evaluate the performance of object proposals, the recall is used in our experiments, 
	which computes the percentage of  ground-truth objects covered by proposals with the IoU value above a threshold.
According to the standard KITTI setup, 
	we set the threshold to $70\%$ for \emph{Car}, and $50\%$ for \emph{Pedestrian} and \emph{Cyclist}. 
We report our experiment results with three recall metrics: 
	Recall vs Number of Proposals with a fixed IoU threshold, 
	Recall vs Various IoU thresholds with a fixed number of proposal, 		and Average Recall (AR) of various IoU thresholds changing from 0.5 to 1 vs Number of Proposals.

\textbf{Evaluation:} 
Since our re-ranking method can be merged to any object proposal generators, 
	we testify its effectiveness on several state-of-the-art baseline generators: EdgeBoxes (EB) \cite{zitnick2014edgeboxes}, DeepBox (DB) \cite{kuo2015deepbox}, Selective Search (SS) \cite{van2011SS}, and 3DOP \cite{Chen20153DOP}. 
Correspondingly, the re-ranked proposals are named as Re-EB, Re-SS, and Re-3DOP, respectively. 
Note that, the re-ranked results of EdgeBoxes and DeepBox are the same, 
	since the DeepBox is the CNN-based re-ranked result of EdgeBoxes.

Figure \ref{fig:recall-proposal} plots Recall vs Number of Proposals with IoU = 0.7 for Cars, and IoU = 0.5 for \emph{Pedestrian} and \emph{Cyclist}.
We can see that in all cases the re-ranked approaches have a visible improvement over original methods. 
The superiority is more obvious especially with a small number of proposals, 
	which indicates that the re-ranked proposals are more effective. 
Clearly, DeepBox is not suitable for the KITTI dataset.
In particular, Re-EB requires only 1,000 proposals to achieve $80\%$ recall for all three classes in the easy difficulty level. 
Furthermore, Re-3DOP achieves 90\% recall with only 1,000 proposals for \emph{Car} in all three difficulty levels. 
Similar improvement is achieved as for Re-SS.

Next, we plot Recall vs IoU threshold at 500 proposals in Figure \ref{fig:recall_500}.
Results show that our method consistently improves the recall of all generators across all IoU threshold,
	especially  at  strict overlap criteria (e.g. IoU > 0.7).
We can see that DB works well with a loose threshold (e.g. IoU > 0.5) while fails at strict one.
Compared to DB, Re-EB significantly improves recall in all IoU thresholds.
Specifically, Re-3DOP achieves largest AUCs (Areas Under Curve) over all classes and difficulty levels.

AR vs Number of Proposals is shown in Figure \ref{fig:ar-proposal}.
As is expected, our approach achieves higher average recall (AR) than baselines, especially at a small number of proposals. 
Particularly, using 1,000 Re-3DOP proposals gives a higher AR than using 2,000 3DOP proposals for \emph{Car} on moderate difficulty level.

\textbf{Impact on Object Detection:} 
In order to further validate the effectiveness of our approach, 
	we employ the fast R-CNN network proposed in \cite{Chen20153DOP} to estimate object detection performance. 
Table \ref{tabel:dtection AP} reports the average precision (AP) of object detection with different number of proposals.
We can see that, when using only  top 10 proposals, the Re-3DOP leads to an AP of $53.33\%$, while 3DOP leads to $35.73\%$.
In particular, compared to 3DOP obtaining the AP of 88.26\% with as many as 5,000 proposals, Re-3DOP achieves an AP of 88.34\% using only 1,000 proposals, 
	indicating that our approach selects more accurate proposals.
Similarly, Re-EB achieves an AP of $76.81\%$ when using 1,500 proposals, 
	while EB only gives $76.62\%$ even using 5,000 proposals.
DB fails to improve detection performance in such strict context. 
As expected, Re-SS gives similar improvement, Re-SS achieves 51.51\% using only 500 proposals, while SS requires 2,000 proposals to obtain 51.43\%.

\textbf{Visualization:}
Figure \ref{fig:exapmle} shows examples of top scoring 100 proposals of 3DOP and Re-3DOP on KITTI dataset.
As can be seen from the figure \ref{fig:exapmle}, 
	our method successfully prunes away false positive proposals, 			while 3DOP includes a lot of irrelevant proposals in the top 100 proposals.

\textbf{Running Time:}
Tabel \ref{tabel:run times} shows running time of each step in  our approach. 
Our approach takes $0.79s$ in total on a singe core. 
Parallel computation can further enables our approach to be real-time.

\section{Discussion and Conclusion}
We have presented a simple and effective class-specific re-ranking approach to improve the recall performance of object proposals in the context of automatic driving. 
We take advantage of semantic segmentation, stereo information, contextual information, CNN-based objectness, and low-level cue to re-score object proposals. 
Experiments on KITTI detection benchmark show that our approach significantly improves the recall rate of object proposals across various IoU threshold. 
Furthermore, we achieve the best recall performance in all recall metrics by merging to 3DOP.
Evaluation on object detection shows that our approach can achieve an higher AP with less proposals.

\section{Acknowledgements}
This work is supported by the Nature Science Foundation of China (No.61202143, No. 61572409), the Natural Science Foundation of Fujian Province (No.2013J05100) and Fujian Provi-nce 2011 Collaborative Innovation Center of TCM Health Management.



\small
\setlength{\bibsep}{1ex}
\bibliography{nips}

\end{document}